\documentclass[letterpaper, 10 pt, conference]{ieeeconf}  

\usepackage[pscoord]{eso-pic}

\IEEEoverridecommandlockouts                
\usepackage{epsfig} 
\usepackage{amsmath} 
\usepackage{amssymb}  
\graphicspath{{./Figures/}}
 \usepackage{subfigure}
\usepackage{algorithm}
\usepackage{algpseudocode}
\usepackage{amsmath}
\usepackage{etoolbox}
\makeatletter
\patchcmd{\@makecaption}
  {\scshape}
  {}
  {}
  {}
\makeatother
\usepackage[pdfusetitle]{hyperref}

\title{Enhanced Visual Feedback with Decoupled Viewpoint Control in Immersive Humanoid Robot Teleoperation using SLAM}
\author{Yang Chen$^*$, Leyuan Sun$^*$, Mehdi Benallegue, Rafael Cisneros-Limón, Rohan P. Singh, Kenji Kaneko,\\ Arnaud Tanguy, Guillaume Caron, Kenji Suzuki, Abderrahmane Kheddar, and Fumio Kanehiro
\thanks{Yang Chen$^{*}$ and Leyuan Sun$^{*}$ are co-first authors and contributed equally to this work. Corresponding author: Yang Chen.}
\thanks{Y. Chen is with the School of Integrative and Global Majors (SIGMA), University of Tsukuba, Japan.
        {\tt\small chenyang@ai.iit.tsukuba.ac.jp}} 
\thanks{L. Sun, R.P. Singh and F. Kanehiro are with Department of Intelligent and Mechanical Interaction Systems, Graduate School of Science and Technology, University of Tsukuba, Japan. }%
\thanks{G. Caron is with Universite de Picardie Jules Verne, MIS lab, Amiens, France.} 
\thanks{All authors are with  CNRS-AIST JRL (Joint Robotics Laboratory), IRL, National Institute of Advanced Industrial Science and Technology (AIST).
        {\tt\small son.leyuansun, mehdi.benallegue, rafael.cisneros, rohan-singh, k.kaneko, guillaume.caron, f-kanehiro @aist.go.jp; arnaud.tanguy, kheddar@lirmm.fr}}%
\thanks{K. Suzuki is with the Faculty of Engineering and Center for Cybernetics Research, University of Tsukuba, Japan. 
        {\tt\small kenji@ieee.org}} 
}

\begin{document}

\maketitle

\begin{abstract}
In immersive humanoid robot teleoperation, there are three main shortcomings that can alter the transparency of the visual feedback: (i) the lag between the motion of the operator’s and robot’s head due to network communication delays or slow robot joint motion. This latency could cause a noticeable delay in the visual feedback, which jeopardizes the embodiment quality, can cause dizziness, and affects the interactivity resulting in operator frequent motion pauses for the visual feedback to settle; (ii) the mismatch between the camera's and the headset's field-of-views (FOV), the former having generally a lower FOV; and (iii) a mismatch between human's and robot's range of motions of the neck, the latter being also generally lower. In order to leverage these drawbacks, we developed a decoupled viewpoint control solution for a humanoid platform which allows visual feedback with low-latency and artificially increases the camera's FOV range to match that of the operator's headset. Our novel solution uses SLAM technology to enhance the visual feedback from a reconstructed mesh, complementing the areas that are not covered by the visual feedback from the robot. The visual feedback is presented as a point cloud in real-time to the operator. As a result, the operator is fed with real-time vision from the robot’s head orientation by observing the pose of the point cloud. Balancing this kind of awareness and immersion is important in virtual reality based teleoperation, considering the safety and robustness of the control system. An experiment shows that the effectiveness of our solution.
\end{abstract}


%
\IEEEpeerreviewmaketitle

\section{Introduction}
\label{intro}
Teleoperation technology is getting a renewed attention due to its potential in transferring human's intelligence and actions to a remote location, whose interest is even more relevant due to the current Covid-19 outbreak. ANA Avatar XPrize~\cite{ANAAvatar} is one of the robotics competitions encouraging this trend. 
In teleoperation, a common setup consists of an operator equipped with a master station to control, through network communication, a robot situated at a remote location. For an immersive control of the robot's head motion and receiving a visual feedback from the robot, a wearable head-mounted display (HMD) is the most common choice.
One of the common issues faced in teleoperation is the lag between operator's and robot's head motion due to network communication delays or slow robot joint motion. The visual rendering of the robot's vision sensor (e.g., embedded stereo camera) is delayed.
\begin{figure}[t]
\centering
\includegraphics[width=0.48\textwidth]{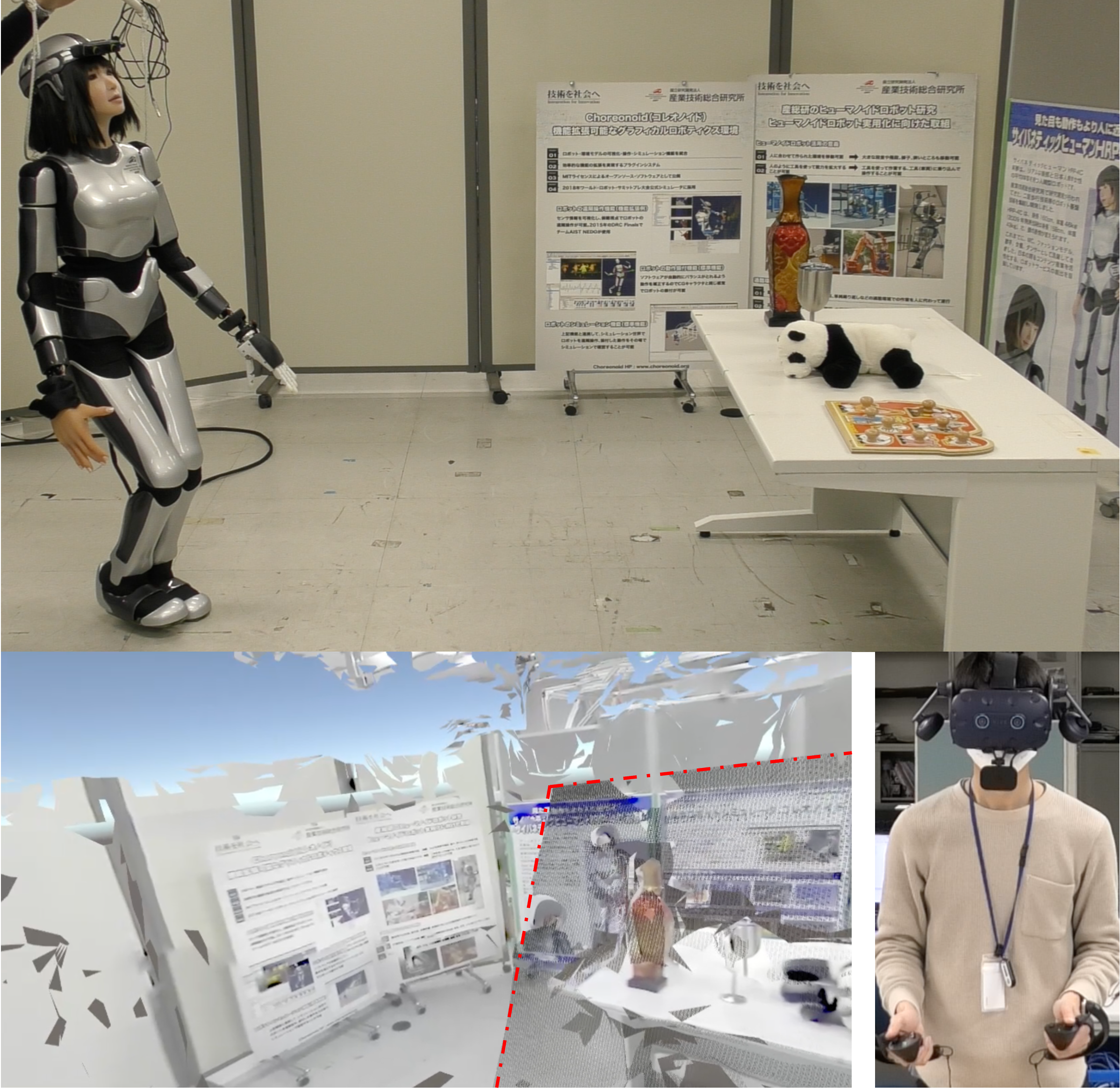}
\caption{Proposed tele-visualization. Top side: real scene, left-down side: virtual scene, the real-time point cloud is shown in the area inside the red dot line, the area outside denotes the constructed mesh, right-down side: operator.}
\label{demo}
\vspace{-0.6cm}
\end{figure}
In an immersive experience, such rendering latency causes dizziness especially when the operator moves her/his head fast \cite{zhao2018effects}. If the viewpoint of HMD is decoupled from the visual information or if its FOV is larger than that of the camera, then there is a blank area that is not covered by the visual feedback from the robot's camera.

Furthermore, humanoid robots often have a lower range of motion w.r.t humans, which may also limit the visual perception if the operator's master station does not constrain the operator head to match the robot's head motion (a highly impractical setup). 

In order to deal with such shortcomings, we propose a decoupled viewpoint solution using SLAM technology for humanoid teleoperation. Our idea is novel and simple to understand: we are using a point cloud as real-time visualization feedback, and a pre-constructed mesh that complement and enhance real-time visual data in blank areas caused by the previously mentioned shortcomings. The latter (tracking latency, robot's joint range limits and camera's smaller FOV) are handled all at once. This is illustrated in Fig. \ref{demo}.

The main contributions of our paper are listed as follows:
\begin{enumerate}
  \item We devised a decoupled viewpoint solution for a humanoid platform which allows the operators to sustain visual feedback changes with low-latency when they freely control their hidden view duplicate in an augmented reality environment.
  \item Use a reconstructed mesh built by SLAM to complement the blank area caused by the mismatch between FOVs, or the mismatch between range of neck joints or the lag between operator's and robot's head motion.
  \item We propose an online calibration solution which could align the virtual HMD and the virtual robot camera frame for the best visualization quality of the point cloud.
\end{enumerate}

\section{Related works}\label{rw}
In this section, we introduce the SLAM technology used in humanoid robots and the different televisualizations in VR teleoperation with a focus on robot application. 

\textit{SLAM in humanoids}: Humanoid robots already use visual SLAM technology for navigation in various environments thanks to their embedded cameras, e.g.~\cite{SLAMinlocomotion}. Recently,~\cite{tsurural} used the map built by SLAM not only for locomotion but also for searching an object and estimating its pose by using point cloud registration.~\cite{Tanguy2016ClosedloopRS} integrated dense SLAM to a QP control framework for localization and also balancing a humanoid robot, which allows using visual odometry to make a reaching plan adjustable online. Besides, SLAM is a fundamental technology when humanoids are used in large-scale manufacturing settings, e.g.,~\cite{RAM}. It is used to provide precise localization to a humanoid within its environment, build a semantic-reconstructed map, define the walking targets and so on. These few examples only show that SLAM is already adopted in several humanoid applications and is part of the humanoid basic planning and control architecture. This work highlights another application of the existing embedded SLAM in the humanoid embodied telepresence context. 

With respect to televisualization works, we introduce a solution using first-person-view for televisualization, which is a common choice to be applied on humanoid platform, as it is more immersive than the third-person-view.

\textit{Stereo RGB fixed}: Displaying real-time stereo RGB image from a stereo camera by means of a VR HMD \cite{advanced_robotics2020,humanoid_vr} is the most common practice in VR teleoperation. Compared with non-VR type, such as video streaming on an external monitor, the HMD allows better immersive capabilities and is more suitable to estimate the depth of objects present in the remote world. However, the latency between the motion of operator's and robot's head induced by a fast motion of the former, compromises the coherency of the visual feedback w.r.t sensory expectations and may cause dizziness. Moreover, this problem compromises intuitive interaction, as explained in Section~\ref{intro}.

\textit{Decoupled viewpoint}: Instead of directly mapping the stereo RGB image to the operator both eyes, decoupled viewpoint means that the operator could realize the an independent motion from the real-time robot's head motion in VR visualization. For example, when the operator moves the head quickly, the direct mapping solution does not update visual feedback instantly, yet a naive \textit{decoupled viewpoint} will generate blank (empty) spaces in the image first and the RGB or point cloud will move towards that space gradually with the speed of the robot's head movement. Theoretically, this solution could reduce the dizziness compared with the direct mapping solution, since the operator could realize that the head motion changes the visual feedback, even if there is only a blank area being displayed. Yet, the latency in visual feedback display in the HMD still exists. The most recent work we could find which applied a decoupled viewpoint control is reported in~\cite{nimbo_avatar,nimbro_spherical}, and is developed by Team NimbRo in the frame of ANA Avatar XPrize. Thanks to the wide-angle camera that they use and the 6D DOF of their robot's head, their solution could allow the operator to freely control their perception in the VR device and the blank space caused by a \textit{decoupled viewpoint} could be complemented by a spherical rendering method that they proposed. However, the limitation of spherical rendering is that the entire scene is considered with a constant depth, or otherwise, a distortion of image happens (as they reported).  

\textit{Decoupled viewpoint with reconstructed mesh or CAD model}: The closest solution we found related to this concept is from Team I-BOTICS\footnote{\url{https://www.youtube.com/watch?v=mL9SyGfuqI4}} for ANA Avatar XPrize using RGB with a CAD model as televisualization feedback. The RGB image provides real-time visual feedback, while the CAD model of the scene complements the latency that causes the aforementioned blank area. Different from this solution, we rather use a point cloud to fuse with the reconstructed mesh built by using SLAM technology. The advantages of our solution are that i) we do not need the knowledge of the environment since the mesh can be reconstructed on-line when the robot explores the working area; and ii) the visualization of point cloud fused (aligned) with the mesh has better immersion properties compared to RGB with CAD and mesh; see Section~\ref{concept}.

For other mobile platforms, a third-person-view is used. For example, in~\cite{iros2019_mobile} the operator can adjust the pose of the viewpoint manually; they also provide a constructed mesh for visualization. However, there is no real-time information to complement the mesh, and the mesh is updated with a low frequency, making it difficult to handle dynamic scene. An intuitive interface for bimanual robot teleoperation system was developed in~\cite{humancontrolview}. RGB merged with 3D point cloud is used for televisualization. The system could track the operator's head motion to make him freely look around using the VR view. However, the limitation of this solution is the need for additional operator support to select these views' positions.

For teleoperation of industrial robotic arms, the third-person-view is the best choice, as a clear and complete observation is more important than immersion.~\cite{robotarmicar} proposed three viewpoints for the operator to select during teleoperation. In the HMD, the operator could select freely to see the model of robot arm from side, front, and top view. Another approach called Picture-in-Picture (PiP) is also common, where video streams from multiple views are displayed concurrently and placed next to each other~\cite{lin2017outside,whitney2020comparing}. Both solutions increase the cognitive burden for the operator.~\cite{multiviewral} provided a multi-visual source merging solution where the local point cloud is projected to a global stereo image nicely, providing additional information for grasping tasks. However, in contrast to our solution, it requires an additional visual sensor. In~\cite{pointcloudwithartificial}, the authors use a VR environment to compensate the difference of FOV between HMD and camera. It has better immersion than not using anything but the artificial environment cannot represent the true surroundings like our reconstructed mesh. So far, the early work of a total VR-based decoupling is reported in~\cite{kheddar2001smc}, but this solution cuts fully the operator from the real perception even if features such as the manipulated objects are updated from the real world.

\section{Methodology}

\subsection{Robot Platform and VR Equipment} \label{rpve}
We use the HRP-4CR robot \cite{janus}, an adult-size humanoid with realistic appearance, which is an upgraded and "refreshed" version of HRP-4C in~\cite{HRP-4C}. A ZED Mini camera (from Stereolabs Inc.) is mounted on the forehead of the robot. On the operator side, an HTC Vive Pro Eye HMD (from HTC Corporation) is worn by the operator to receive the visual information and transfer the control command to the robot's head at the same time. The joystick which is mounted on a Valve Index controller (from Valve Corporation) is used for controlling the locomotion of the robot.

Our method is general enough to apply to any other similar set-up with humanoid, camera and HMD providing few coding adjustments.

\subsection{System architecture}
\begin{figure}[t]
\begin{center}
\includegraphics[width=0.48\textwidth]{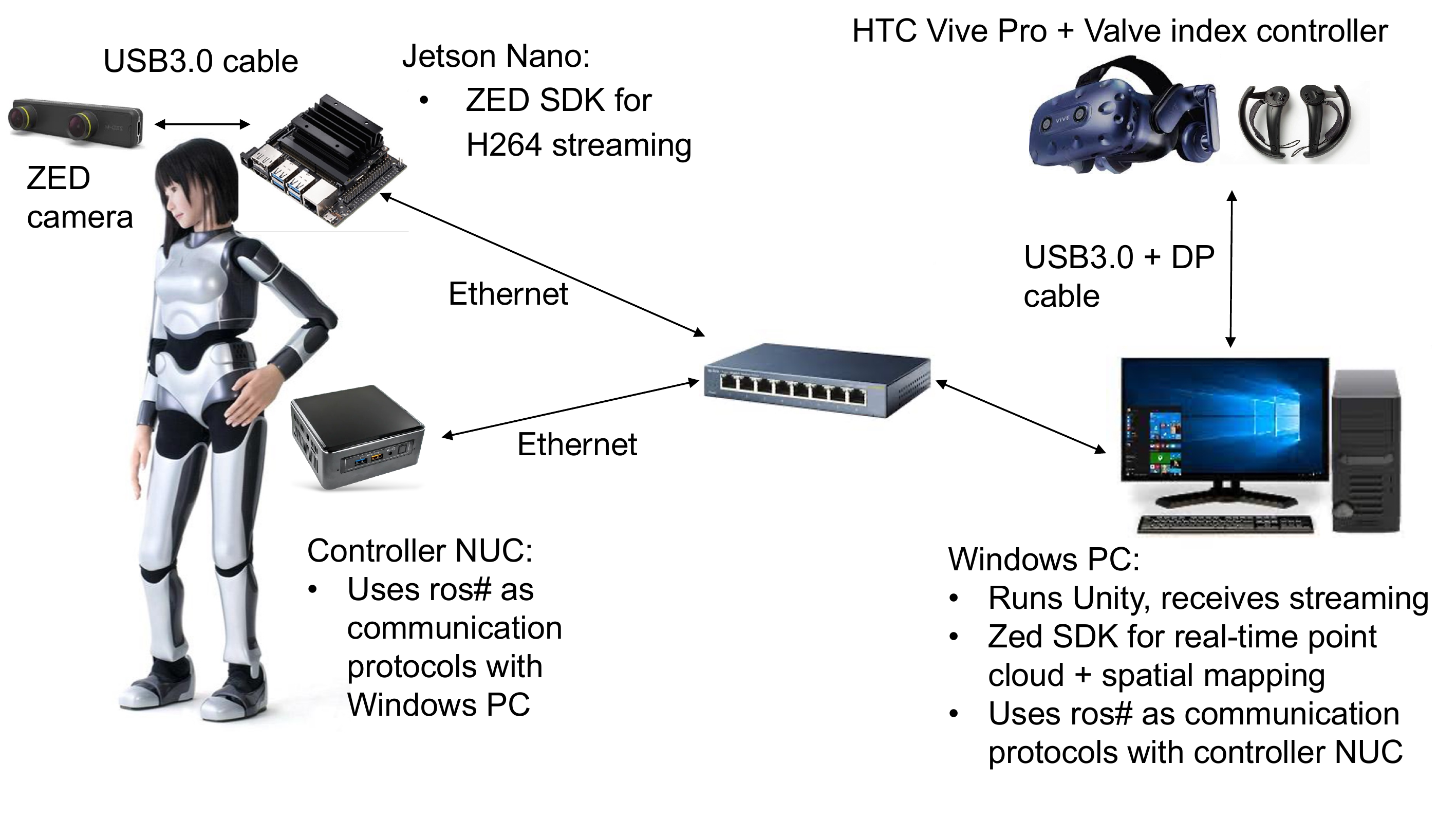}
\caption{System architecture.}
\label{hardware}
\end{center}
\vspace{-0.6cm}
\end{figure}

The robot control unit is an Intel NUC computer, a Jetson Nano (from Nvidia Corporation) is used for interfacing with the ZED Mini camera and streaming the visual information over a wired-LAN. The operator station is driven by a PC with specification (Intel i9-9900K 3.6~GHz, Nvidia RTX 2080Ti), mainly for running Unity3D (the platform we used for building a virtual space). All computers are connected by Ethernet. The system architecture is shown in Fig.~\ref{hardware}. The communication protocol between NUC and Windows PC is ROS-Sharp\footnote{\url{https://github.com/siemens/ros-sharp}}. 

\subsection{Conceptual evaluation of visual expressions}\label{concept}
\begin{figure}[t]
\begin{center}
\includegraphics[width=0.45\textwidth]{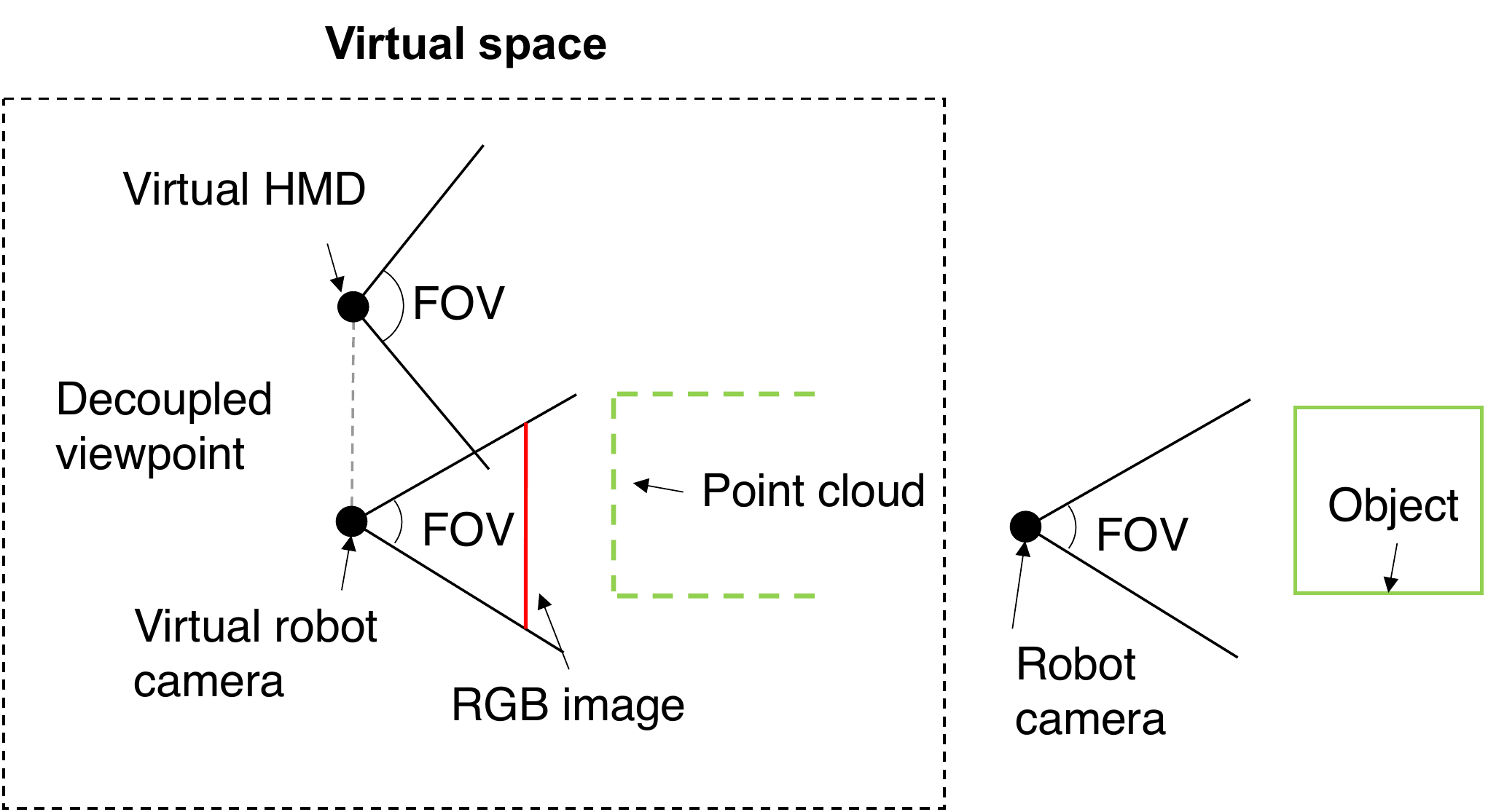}
\caption{Concept analysis. The virtual HMD should be able to see what the real robot camera can see, in the scenario denoted above, the distortion could happen. }
\label{2d_concept}
\end{center}
\vspace{-0.6cm}
\end{figure}

Compared to the method that fuses 3D mesh and 2D RGB image, fusing 3D point cloud and 3D mesh would result (in theory) in a better merging visual effect. As a result, one expects a better immersion for the operator. The difficulty of fusing 2D RGB image and 3D mesh comes from the distortion of a 2D image when observed from different viewpoints.
A concept of visualization is illustrated in Fig.~\ref{2d_concept}, where there is a decoupled movement between the virtual HMD camera and the virtual robot camera in a VR environment (virtual space). Consider that the robot camera is capturing a 3D object such as a cube, then a RGB image or point cloud can be generated in the virtual sapce, due to the 2D property of the RGB image, the distortion can happen when it is observed from the virtual HMD. The distortion can be much less if 3D point cloud are used instead and if we assume the shape of point cloud is similar to the object itself. 

The fusion quality between 3D point cloud and 3D mesh relies on the quality of the constructed mesh, the point cloud and the accuracy of odometry. Here we use a state-of-art 3D construction method called "Spatial Mapping" \footnote{\url{https://www.stereolabs.com/docs/spatial-mapping/}} and ZED visual-inertial odometry. With the odometry as a memory of the point cloud localization, the alignment between the constructed mesh and point cloud can be accomplished automatically. An expected fusion-effect example is shown in Fig.~\ref{demo}; compare the real scene (top side) and the virtual scene (left-down side), there we see that the mesh is well-fused with the real-time point cloud. 

A fusion between the constructed mesh and the real-time point-cloud with an additional decoupled viewpoint control method will be presented in Section \ref{construction}. Using that method, the operator is able to intuitively control the perspective to observe the environment as s/he wants, increasing the immersion. However, in our tele-operation scenario, being aware of the difference between self and robot is important, since there exists a significant difference between a human and the avatar robot. Considering that the operator moves the head quickly, the perspective also changes fast. If the operator is not aware that the robot head motion is delayed, then the operation may cause an accident, e.g. a robot collision with the environment.
Our proposed solution is to make the color of the constructed mesh a bit different from the real color of the environment. Therefore, the operator can distinguish the mesh and real-time point-cloud to achieve a balance between awareness and immersion. This is also helpful when the environment is dynamic, as the operator would put more trust on the real-time information instead of the mesh.

\subsection{Construction of virtual space \& teleoperation mapping}\label{construction}

 \begin{figure*}[t]
	\centerline{\includegraphics[width=0.8\textwidth]{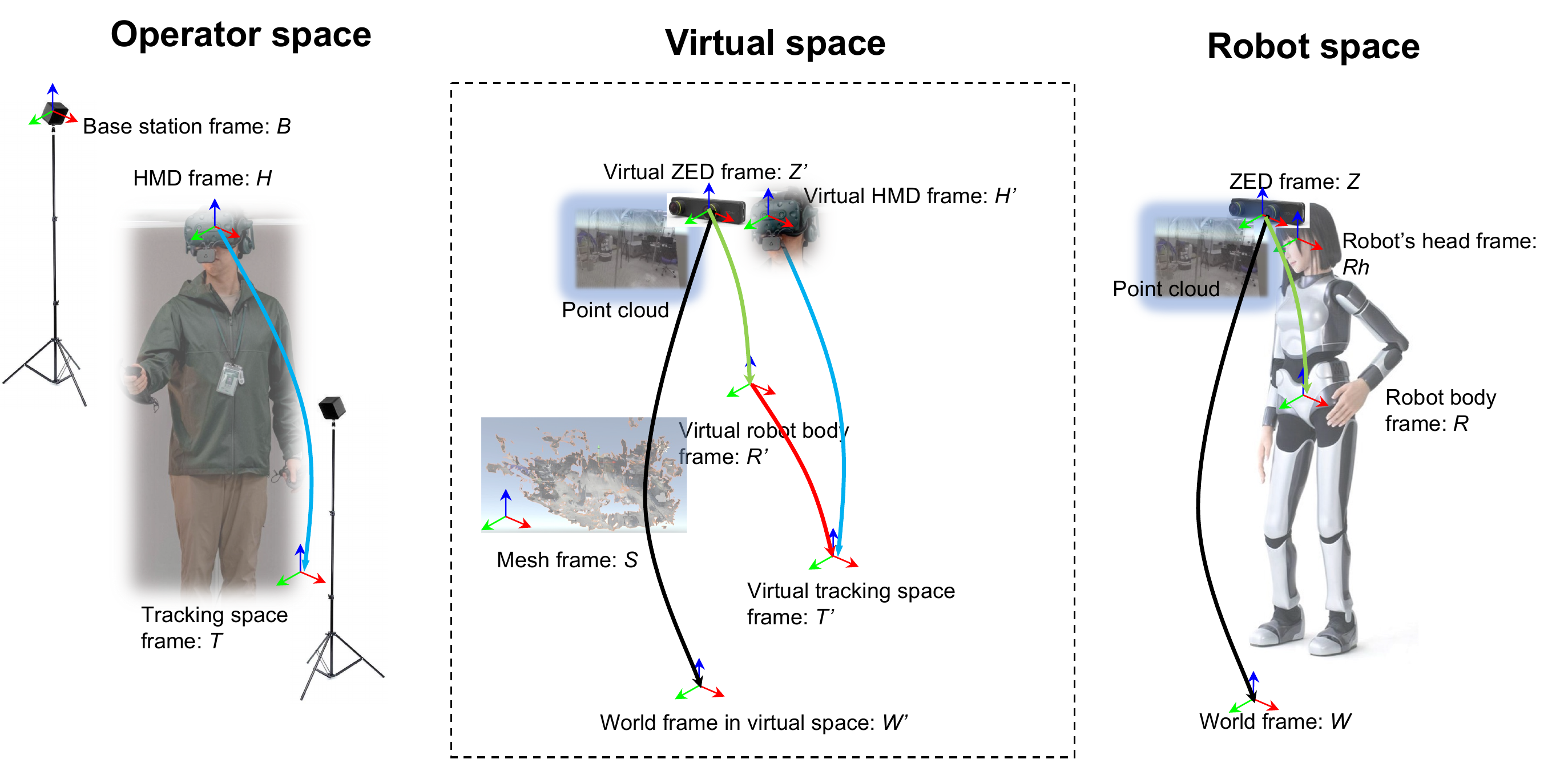}}
	\caption{Teleportation mapping: the motion of the operator's head, robot's head and ZED camera are mapped to the virtual space as depicted by the blue, green and black lines.}
	\label{mapping_0}
	\vspace{-0.3cm}
\end{figure*}

The creation of a virtual space is necessary for the VR televisualization system which we introduced in Section \ref{rw}. From the solution of \textit{Stereo RGB fixed} to \textit{decoupled viewpoint with reconstructed mesh or CAD model}, the complexity of the virtual space increases. Here we explain the key difference between those solutions; specifically, we introduce the construction method of our proposal, which lies on the category of \textit{decoupled viewpoint with reconstructed mesh or CAD model}.

For the \textit{stereo RGB fixed}, the virtual space is the simplest situation: map the image of each camera eye to each eye of the HMD. For the \textit{decoupled viewpoint}, we need to represent the decoupled motion of a robot camera and HMD in the virtual space. For the \textit{decoupled viewpoint with reconstructed mesh or CAD model}, because we have the mesh as a world reference in the virtual space, not only the relative relation between robot camera and HMD has to be represented properly, but also the relation between robot camera, HMD and the mesh.

To better demonstrate the mapping relationship in our tele-visualization system, we have separated the space into three parts: operator space, virtual space, and robot space, as shown in Fig. \ref{mapping_0}. All frames are described in TABLE \ref{tableauc}.
    \begin{table}[htb]
    \renewcommand\arraystretch{1.2}
    \newcommand{\tabincell}[2]{\begin{tabular}{@{}#1@{}}#2\end{tabular}}
    \caption{Frame definition} \label{tableauc}
    \begin{center}
    \resizebox{8.7cm}{!}
    {
    \begin{tabular}{cccccccc}
    \hline
    \multicolumn{1}{l|}{\textbf{Virtual space}} & \textbf{Operator and Robot space}         \\ \hline
    Virtual HMD frame: \textit{\textbf{H'}}                      & HMD frame: \textit{\textbf{H}}              \\
    Virtual ZED frame: \textit{\textbf{Z'}}              & ZED frame: \textit{\textbf{Z}}         \\
   Virtual tracking space frame: \textit{\textbf{T'}}           & Tracking space frame: \textit{\textbf{T}} \\
   Virtual robot body frame: \textit{\textbf{R'}} &
   Robot body frame: \textit{\textbf{R}}              \\
    World frame in virtual space: \textit{\textbf{W'}}       &  World frame: \textit{\textbf{W}}     \\
    Mesh frame: \textit{\textbf{S}} &
    Robot's head frame: \textit{\textbf{Rh}} \\
    Base station frame: \textit{\textbf{B}}
    \end{tabular}
    }
    \end{center}
    \end{table}
    
The measurements are shown in Eq. \ref{equation1}: The description of the HMD frame in the tracking space frame is $M_{H}^{T} = M_{B}^{T} \cdot M_{H}^{B}$, where $M_{B}^{T}$ is a fixed value calibrated when we did the VR room setup procedure, $M_{H}^{B}$ is a real-time measurement made by the base station (from Valve Corporation), and $M_{H}^{T}$ represents the head movement of the operator. The ZED pose relative to the robot body frame is $M_{Z}^{R} = M_{Rh}^{R} \cdot M_{Z}^{Rh}$, which are measured from the forward kinematics and the extrinsic respectively. $M_{Z}^{W}$ is measured by the ZED visual-inertial odometry in world frame (using ZED SDK). Then,

\begin{equation}	
\mathbf{M} = [M_{H}^{T},M_{Z}^{R},M_{Z}^{W}]
	\label{equation1}
\end{equation}	

Those ones are also mapped from the operator and robot space to the virtual space as shown below:
\begin{equation}
\begin{aligned}
\mathbf{M^{'}} &\stackrel{\text{def}}{=} \mathbf{M},\\
\mathbf{M^{'}} &= [M_{H^{'}}^{T^{'}},M_{Z^{'}}^{R^{'}},M_{Z^{'}}^{W^{'}}]. 
	\label{equation2}
\end{aligned}
\end{equation}



In order to realize the decoupled motion between the virtual ZED frame and the virtual HMD frame, we rely on Eq. \ref{equation2}, namely

\begin{equation}
\begin{aligned}
M_{Z^{'}}^{H^{'}} &= M_{T^{'}}^{H^{'}} \cdot M_{R^{'}}^{T^{'}} \cdot M_{Z^{'}}^{R^{'}} \\
                  &= M_{T}^{H} \cdot M_{R^{'}}^{T^{'}} \cdot M_{Z}^{R},
	\label{equation2}
\end{aligned}
\end{equation}
where $M_{R^{'}}^{T^{'}}$ is a static transformation which can be calibrated in order to display point cloud in HMD properly,. The details of the calibration will be discussed in Section \ref{calibration}.
When there is no mesh, $M_{Z^{'}}^{W^{'}} = M_{R^{'}}^{W^{'}} \cdot M_{Z^{'}}^{R^{'}}$, $M_{T^{'}}^{W^{'}} = M_{R^{'}}^{W^{'}} \cdot M_{T^{'}}^{R^{'}}$, and $M_{H^{'}}^{W^{'}} = M_{R^{'}}^{W^{'}} \cdot M_{T^{'}}^{R^{'}}\cdot M_{H^{'}}^{T^{'}}$ can be determined by any given constant value of $M_{R^{'}}^{W^{'}}$. As a result, we realize a flow: the operator moves the head $\rightarrow$ the robot's head follows this movement with a motion delay $\rightarrow$ the point cloud in virtual space moves together with the ZED frame with a network delay $\rightarrow$ the visual feedback described in the HMD frame updates accordingly. 

However, when there is a mesh, to align point cloud and mesh properly, instead of being a constant value, $M_{R^{'}}^{W^{'}}$ should be the odometry of robot's body frame w.r.t. the world frame. Here, we rely on the ZED visual-inertial odometry, and use Eq. \ref{equation2_0} to realize
\begin{equation}	
    M_{R^{'}}^{W^{'}} \stackrel{\text{def}}{=} M_{R}^{W} = M_{Z}^{W}\cdot {M_{Z}^{R}}^{-1}.  
	\label{equation2_0}
\end{equation}

The precision of ZED visual-inertial odometry is studied in \cite{alapetite2020real}. $M_{R}^{W}$ can also be obtained by other localization methods such as motion capture or the kinematics-inertial odometry of the robot itself, but those methods either require additional sensor or they are not precise enough for our requirements. In the case of having a mesh, we have another flow for locomotion: the operator controls the robot's locomotion by joystick $\rightarrow$ the robot's body moves with a motion delay $\rightarrow$ the point cloud in virtual space moves together with the ZED frame with a network delay $\rightarrow$ the visual feedback described in the HMD frame updates accordingly. 


\subsection{Online calibration}\label{calibration}

\begin{figure}[t]
\begin{center}
\includegraphics[width=0.4\textwidth]{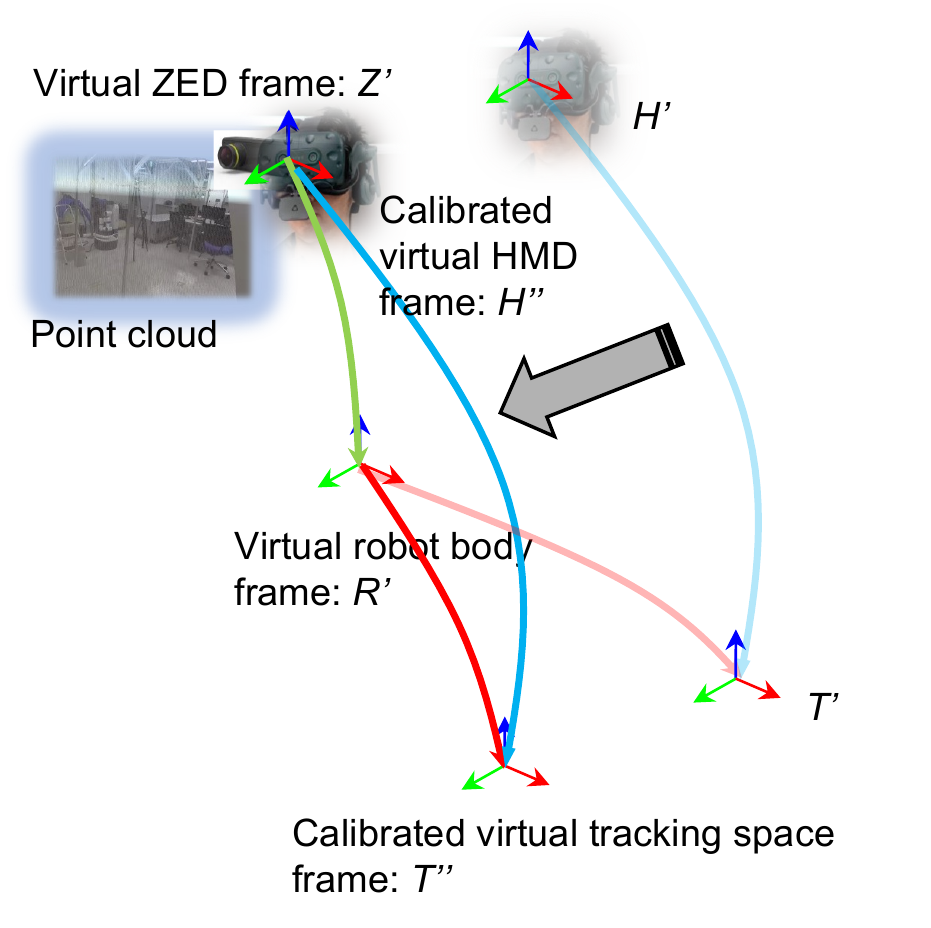}
\caption{Calibration process for the best viewpoint of the point cloud visualization: move the virtual tracking space frame from $T'$ to $T''$ in order to align $H''$ with $Z'$. $H''$ and $T''$ are the calibrated virtual HMD frame and calibrated virtual tracking space frame during the calibration process. After the calibration, $H''$ coincides with $Z'$.}
\label{calibration}
\end{center}
\vspace{-.5cm}
\end{figure}

\begin{figure}[hbt!]
\begin{center}
\includegraphics[width=0.49\textwidth]{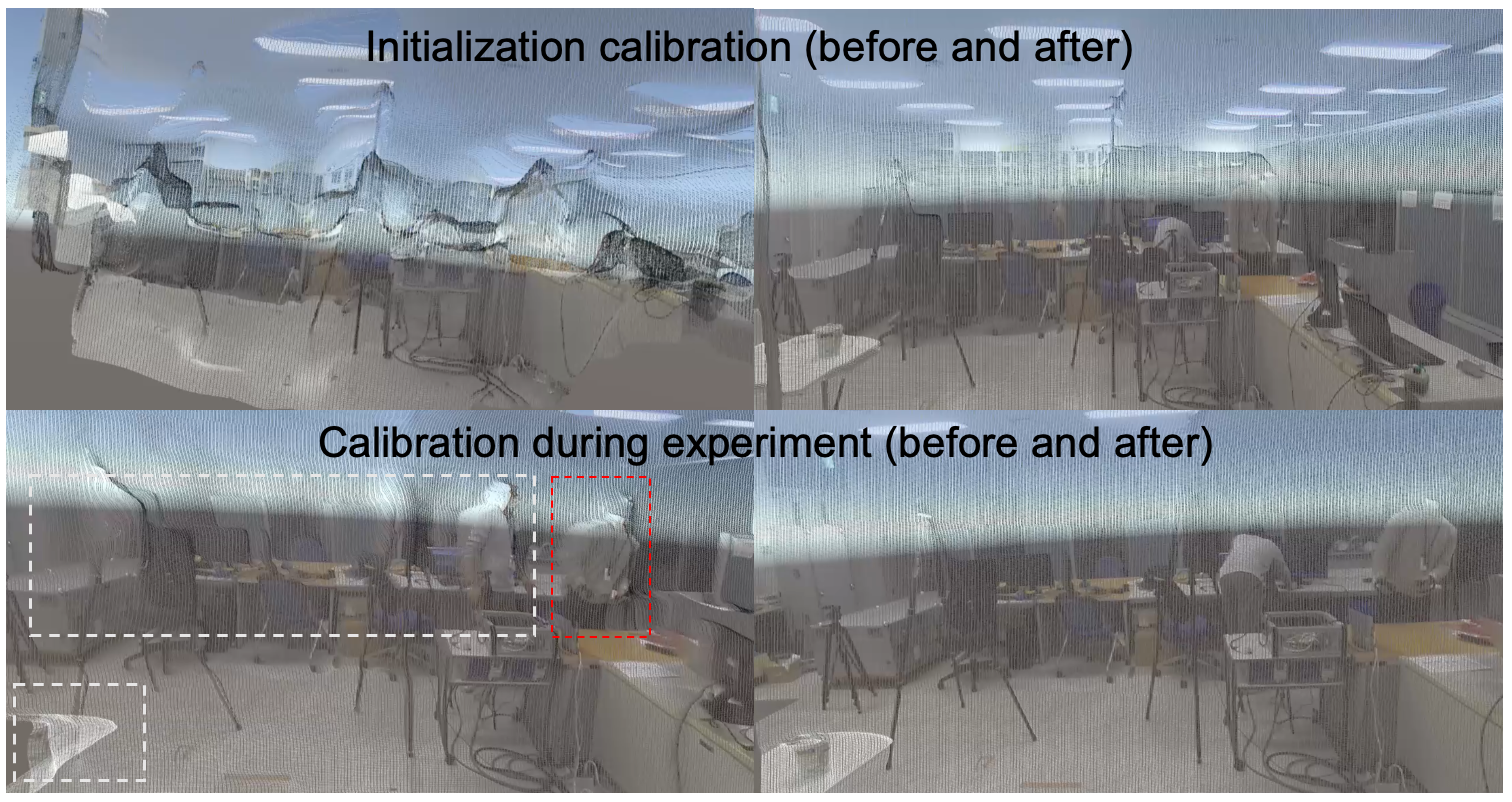}
\caption{Top: initialization calibration, bottom: calibration during experiment, distortion shown in white rectangle caused by the operator shifting position as shown in the red rectangle. Left represents before, right represents after calibration.}
\label{calibration_demo}
\end{center}
\vspace{-0.3cm}
\end{figure}

We have realized the decoupled motion control for both virtual ZED frame and virtual HMD frame by using Eq. \ref{equation2}; however, the height of users can be different, or the user may also move the body slightly during operation. Both factors cause a change of $M_{H^{'}}^{T^{'}}$, resulting in the virtual HMD frame misalignment with respect to the virtual ZED frame. The visualization of point cloud from different viewpoints should have a similar effect as we analyzed in Section \ref{concept}. However, as the point cloud is usually not generated perfectly, its visualization from an offset viewpoint might cause a distortion as shown in Fig. \ref{calibration_demo}. Furthermore, the misalignment can cause the operator to misjudge the relative pose between objects and robot.
In order to visualize the point cloud correctly inside the FOV of the HMD, we propose the online calibration method shown in Fig. \ref{calibration}. 
The calibration problem is defined as bellow:

\begin{equation}
\begin{aligned}
M_{H^{''}}^{W^{'}} &= M_{Z^{'}}^{W^{'}}, \\
\textrm{s.t.} \quad  M_{T^{''}}^{H^{''}} &= M_{T^{'}}^{H^{'}},  \\
	\label{equation6}
\end{aligned}
\end{equation}
where $H^{''}$ and $T^{''}$ represent the calibrated virtual HMD frame and calibrated virtual tracking space frame. The relative pose between the virtual HMD frame and the virtual tracking space frame is always decided by the localization based on the base station's measurement; so instead of changing the pose of the virtual HMD frame directly, we move the virtual tracking space frame from $T^{'}$ to $T^{''}$ to align the calibrated virtual HMD frame with the virtual ZED frame as decomposed by Eq. \ref{equation8}:

\begin{equation}	
\begin{aligned}
    M_{H^{''}}^{W^{'}} &= M_{R^{'}}^{W^{'}}  \cdot M_{T^{''}}^{R^{'}} \cdot M_{H^{''}}^{T^{''}} \\
                       &= M_{Z}^{W}\cdot {M_{Z}^{R}}^{-1} \cdot M_{T^{''}}^{R^{'}} \cdot M_{H}^{T}. \\
	\label{equation8}
\end{aligned}
\end{equation}



We can obtain transformation from the $T''$ to $R'$ by using Eq.\ref{equation7}, which is the one  we need to update the alignment:

\begin{equation}
\begin{aligned}
M_{T^{''}}^{R^{'}} &=  M_{Z^{'}}^{R^{'}} \cdot M_{T^{''}}^{Z^{'}} \\&=
 M_{Z}^{R} \cdot M_{T^{''}}^{H^{''}} \\&=
 M_{Z}^{R} \cdot M_{T}^{H}. 
	\label{equation7}
\end{aligned}
\end{equation}

One example of the effect of calibration is shown in Fig. \ref{calibration_demo}. There are also more things to note: (1) The calibration can be activated by the operator voluntarily pressing a trigger on the controller during tele-operation. (2) The calibration usually only needs to be done when the operator has changed or when the operator's position changes too much during the tele-operation process, as most of the time the point cloud looks good enough even with a small misalignment. (3) The calibration can be activated at any moment and can finish instantly during the operation. (4) It's better to calibrate with a looking forward posture and in a static status, with consideration for the best effect on the decoupled viewpoint.


\begin{figure}[!t]
\begin{center}
\includegraphics[width=0.48\textwidth]{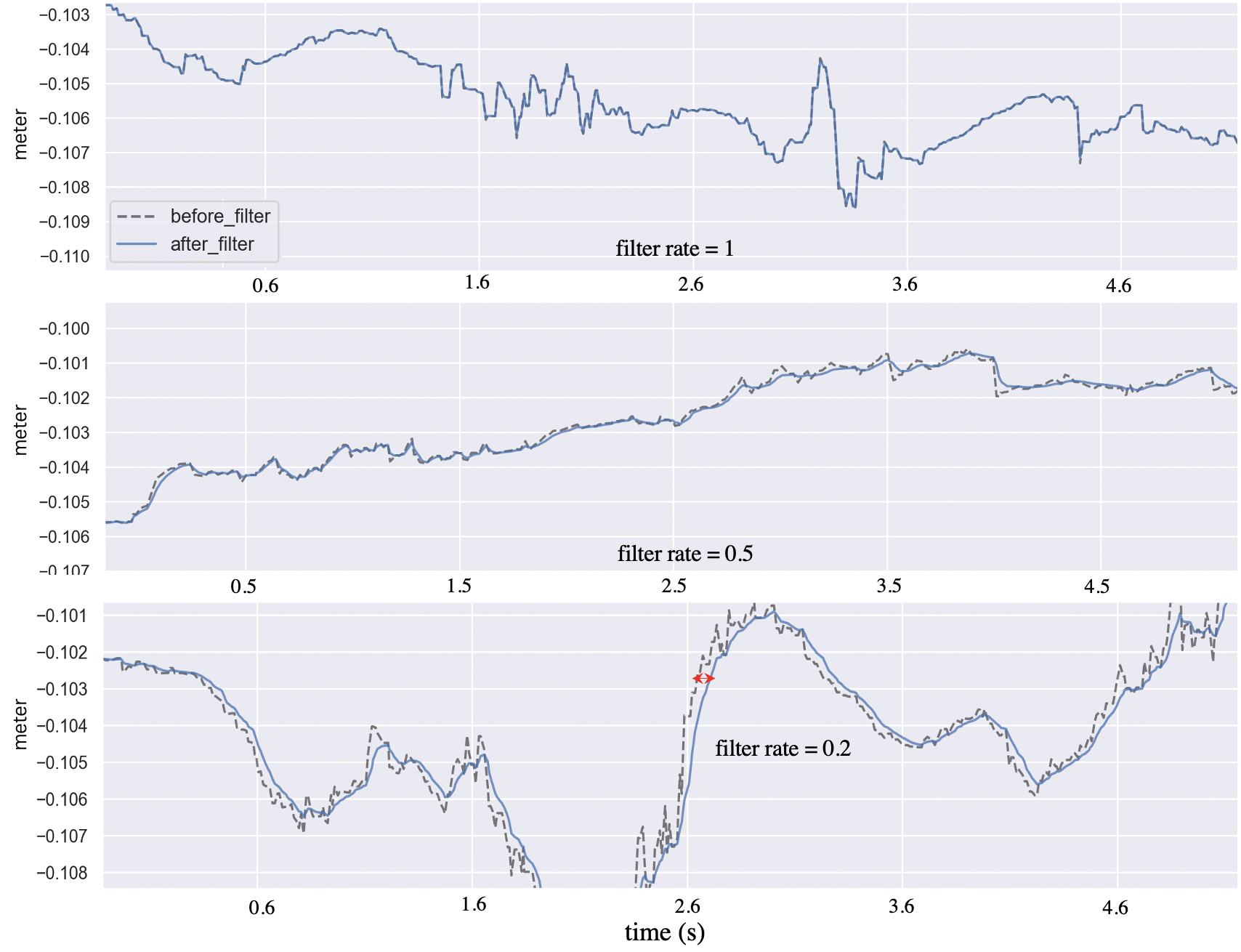}
\caption{The latency caused by different rates of the low-pass filter.}
\label{filterlatency}
\end{center}
\vspace{-0.6cm}
\end{figure}

\subsection{Low Pass Filter}\label{lpf}
Like other related works \cite{nimbo_avatar} \cite{nimbro_spherical} and \cite{iros2019_mobile}, we also need to apply a low-pass filter in order to increase the smoothness of motion and immersion of VR experience. Before its implementation, we first analyze the source of noise.
The noise resulting from observing the point cloud from the HMD could be analyzed by using Eq. \ref{equation2}, where all three values are stable measurements. $M_{T}^{H}$ and $M_{R^{'}}^{T^{'}}$ are local measurements on the Unity side, while $M_{C}^{Z}$ is measured on the controller PC and sent by Rosbridge communication protocol through Ethernet to the Unity side. The noise arises mostly from the last one due to the possible instability of Rosbridge communication and the network.

The noise resulting from observing the mesh from the HMD can be analyzed by Eq. \ref{equation10}. Compare it with Eq. \ref{equation2}. Here, $M_{S}^{W^{'}}$ is a static measurement on the Unity side, $M_{W}^{Z}$ is the ZED odometry measurement sent through network but not by Rosbridge. The noise resulting from observing the mesh could be larger than observing the point cloud due to the usage of the ZED odometry. 

\begin{equation}
\begin{aligned}
M_{S}^{H^{'}} &=   M_{W^{'}}^{H^{'}}\cdot M_{S}^{W^{'}} \\&=
M_{T}^{H} \cdot M_{R^{'}}^{T^{'}} \cdot M_{Z}^{R} \cdot M_{W}^{Z} \cdot M_{S}^{W^{'}} 
	\label{equation10}
\end{aligned}
\end{equation}

In order to reduce the dizziness of the operator caused by the noise, we apply a low-pass filter to the virtual tracking space frame.

In Fig. \ref{filterlatency}, we plot the comparison between the measurement before and after applying the filter using different filter rates. We observe that the high frequency component is removed and when we decrease the rate, the curve becomes smoother, while the lag increases. In our experiment, we use a filter rate of 0.2, such that the latency caused by the low-pass filter is approx. 60 ms. In the future, we will use a filter without delay, such as the complementary filter [20], as the performance is more suitable for VR.

\begin{figure}[t]
\begin{center}
\includegraphics[width=0.48\textwidth]{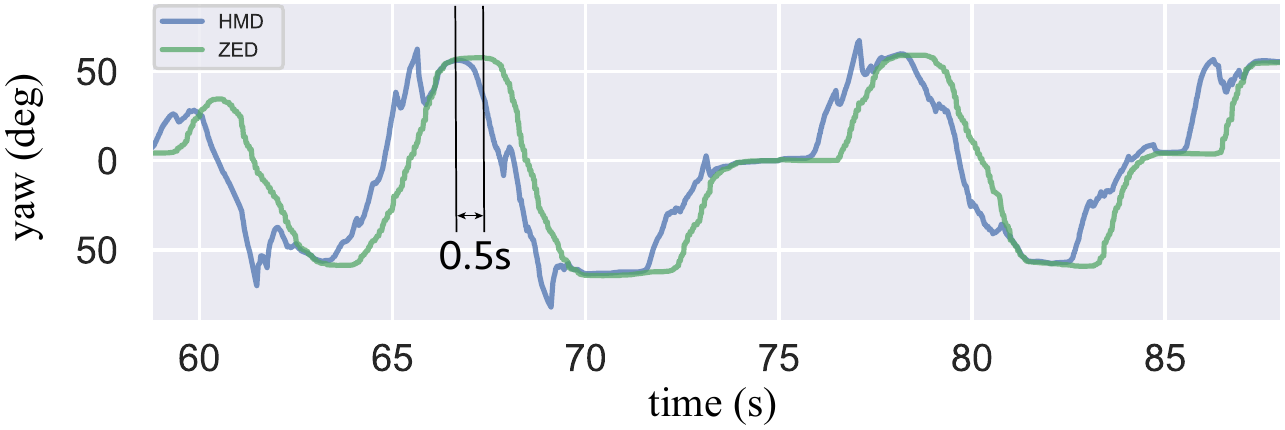}
\caption{Common latency between the motion of operator's and robot's head: approx. 0.5 s.}
\label{latency}
\end{center}
\vspace{-0.3cm}
\end{figure}

\begin{figure}[t]
\begin{center}
\includegraphics[width=0.48\textwidth]{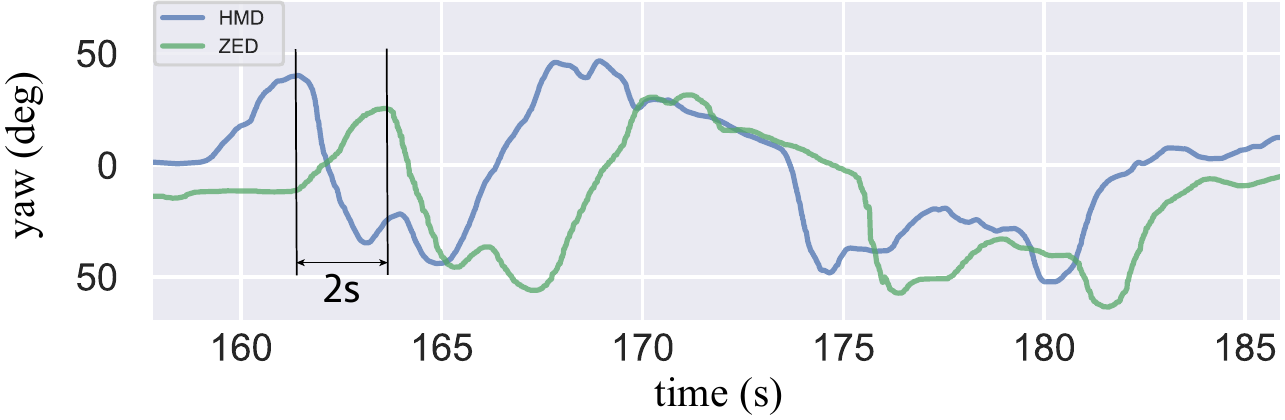}
\caption{Latency when the network was unstable: approx. 2 s.}
\label{highlatency}
\end{center}
\vspace{-0.3cm}
\end{figure}

\begin{figure}[!t]
\begin{center}
\includegraphics[width=0.48\textwidth]{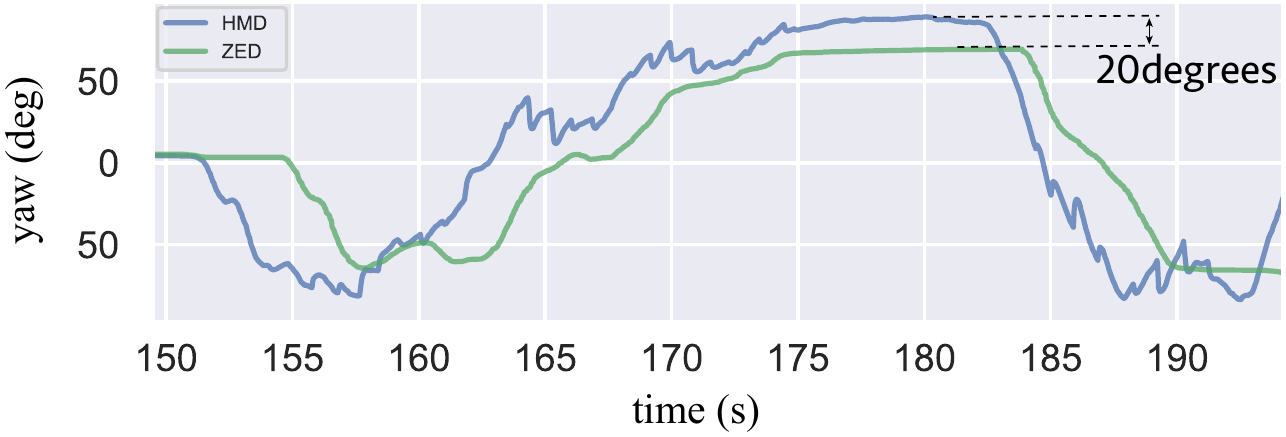}
\caption{Range of motion measurement}
\label{rangeomm}
\end{center}
\vspace{-0.6cm}
\end{figure}

\section{Experiment}

\begin{figure*}[hbt!]
\centering
\includegraphics[width=0.98\textwidth]{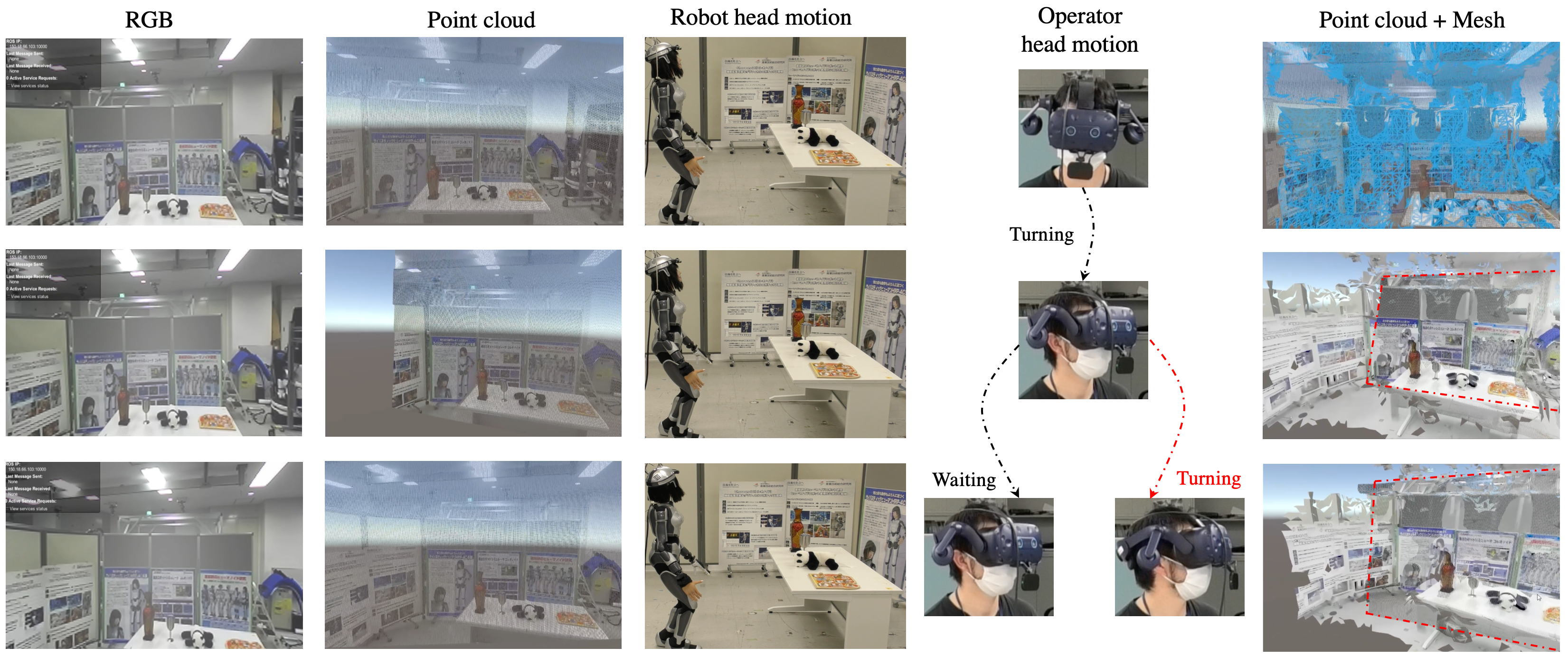}
\caption{Three solutions were compared during the experiment. In the first, second and last columns, we show the image displayed in the HMD for the three different solutions. For the point cloud with mesh (last column, first row), the blue grid shows the user scanned area while the mesh is under construction. In the second and third rows, we marked the real-time point cloud by a red dot line. In the fourth column, from top to down, the displayed images correspond to different operator's motions. In cases of RGB and point cloud, the operator turned to the left from static forward status and then waited a bit for the real-time visual feedback. While in case of the point cloud with mesh, the operator kept turning. The third column shows the robot's head motion, and due to the lag, the robot' head motion does not change in the second row, although the operator had turned his head.}
\label{solutions comparison}
\end{figure*}

\subsection{Latency Measurement}\label{lm}
We measured the main latency in the system between the motion of operator's and robot's head. We could observe these two motions in Unity3D which is the software where our virtual space is built. We ask the operator to turn the head from left to right (approx. -75 deg to 75 deg). In Fig. \ref{latency}, usually when network is stable, we could see the latency is about 0.5 s. However, the latency could increase to 2 s as depicted in Fig. \ref{highlatency} when the network became unstable.

\subsection{Range of Motion Measurement}\label{RoM}
One advantage of our system is that it is able to compensate the blank area caused by the mismatch between the range of motions of human's and robot's heads. The motion of both ZED and HMD are shown in Fig. \ref{rangeomm}. As expected, after the operator approached the maximum angle of head motion, there were 20 deg of gap between the human's head and robot's head in a static situation. In case of the \textit{stereo RGB fixed} solution, the operator would lose this information of the environment. In case of our solution, the operator could see the pre-built mesh. The qualitative result is shown in the next section.

\subsection{Qualitative Evaluation}\label{qe}

We compared our proposed solution with a common \textit{stereo RGB fixed} solution which is adopted by most of teams in the XPRIZE competition such as \cite{snu} \cite{avatrina} as qualitative evaluation. Since our proposal mainly include two parts: one is the decoupled representation in a virtual space, another one is to overlap the point cloud with mesh, we evaluate this two parts separately.
Seven participants (males between their 20s and 40s) were asked to finish one small task three times. The task was to tele-operate the robot to walk straight forward or backward freely and, at the same time, they could turn the head freely to observe the environment as they wanted. For each trial, different visual feedback solutions were provided: stereo fixed viewpoint with RGB image only, decoupled viewpoint with point cloud only or decoupled viewpoint with point cloud and mesh, which correspond to the three solutions that we explained in Section \ref{rw}. The experiment scene is shown in Fig. \ref{solutions comparison}.
For the second and third solutions, the participants were able to calibrate the viewpoint at any time during the operation. For the third solution, the participants were asked to build the mesh right after we started the system, and they could scan the area as much as they wanted. After building the mesh, the point cloud always overlapped the constructed mesh. In order to reduce the burden to the network, we set the point cloud resolution as HD720; however, we also include the performance of point cloud HD1080 in a demo video \footnote{\url{https://www.youtube.com/watch?v=Jdiaosp_qH8}}.

The comparison snapshots of the three solutions are shown in Fig. \ref{solutions comparison}. From the solution of point cloud and point cloud with mesh, we could observe that when the operator turned his head, the real time point cloud could track the center of the HMD display with a delay. In the solution of point cloud with mesh, we could see the fusion between the point cloud and the constructed mesh, as well as that the mismatch between the ZED's and HMD's FOV is complemented by the pre-built mesh. One interesting thing we found is that since the reconstructed mesh could complement the blank area, the operators usually did not wait for the real-time point cloud to come back to the center of the HMD's FOV, but they would keep intuitively moving their perspective, which significantly increased the interaction efficiency with robot. This fact is represented in Fig. \ref{solutions comparison}: after turning the head a small angle, the operator kept turning his head to a larger degree until he saw the blank area which was not covered by mesh (also beyond the motion range of the robot's head) as the last figure shows in the last column.

\section{Conclusion and future work}
In this work, we proposed a balanced-immersive tele-visualization solution with decoupled viewpoint control and complementary visual feedback for humanoid robot teleoperation. The proposed solution mainly contains two core ideas. The first idea is to complement the real-time point cloud with a constructed mesh, aligning them by applying visual-inertial odometry while distinguishing them by changing the color of the mesh, resulting in a balanced fusion effect. The second idea is to realize the decoupled movement between robot's and operator's viewpoints in a virtual space to reduce the visual latency. We achieved a balanced immersion by combining both ideas. We evaluated both ideas by comparing against a standard stereo RGB fixed view solution and verified the effectiveness of our proposal regarding the provision of instant visual feedback, bigger FOV and bigger range of view (due to a bigger range of human head motion). We also identified several limitations that could be improved in the future. One limitation comes from the quality of the reconstructed mesh and the odometry precision leading to a misalignment between point cloud and mesh, which can be solved by a continuously improved higher precision SLAM method, available nowadays. Also, the real-time part should always be given a higher visual priority than the constructed mesh. This could be achieved by using other rendering methods. Another future work will be evaluate the proposed system with a complete user study evaluated by questionnaire. Our proposal is most suitable for a humanoid platform, but potentially it could also be applied to other robotic platforms, especially those using human head motion to control a robot's head with a mounted camera.

\bibliographystyle{IEEEtran}
\bibliography{IEEEabrv,C_SLAM}

%






\end{document}